\pdfoutput=1

\documentclass[11pt]{article}

\usepackage[final]{acl}

\usepackage{times}
\usepackage{latexsym}

\usepackage[T1]{fontenc}

\usepackage[utf8]{inputenc}

\usepackage{microtype}

\usepackage{inconsolata}
\usepackage{amsmath,amssymb,amsfonts}
\usepackage{algorithmic}
\usepackage{graphicx}
\usepackage{subfigure}
\usepackage{textcomp}
\usepackage{xcolor}
\usepackage{enumitem}
\usepackage{bm}
\usepackage{multirow}
\usepackage{booktabs}
\usepackage{graphics}
\usepackage{keyval}
\usepackage{trig}
\usepackage{threeparttable}
\usepackage{marvosym}
\usepackage{newtxmath}

\usepackage{tabularx}

\title{Graceful Forgetting in Generative Language Models}

\author{Chunyang Jiang, Chi-min Chan, Yiyang Cai, Yulong Liu, Wei Xue\textsuperscript{\Letter}, Yike Guo\textsuperscript{\Letter} \\
  Hong Kong University of Science and Technology \\
  \texttt{cjiangaq@connect.ust.hk} \\}

\begin{document}
\maketitle
\renewcommand{\thefootnote}{}
\footnotetext{\textsuperscript{\Letter} The corresponding authors are Wei Xue and Yike Guo.}
\begin{abstract}
Recently, the pretrain-finetune paradigm has become a cornerstone in various deep learning areas. While in general the pre-trained model would promote both effectiveness and efficiency of downstream tasks fine-tuning, studies have shown that not all knowledge acquired during pre-training is beneficial. Some of the knowledge may actually bring detrimental effects to the fine-tuning tasks, which is also known as \textit{negative transfer}. To address this problem, \textit{graceful forgetting} has emerged as a promising approach. The core principle of graceful forgetting is to enhance the learning plasticity of the target task by selectively discarding irrelevant knowledge. However, this approach remains underexplored in the context of generative language models, and it is often challenging to migrate existing forgetting algorithms to these models due to architecture incompatibility. To bridge this gap, in this paper we propose a novel framework, Learning With Forgetting (LWF), to achieve graceful forgetting in generative language models. With Fisher Information Matrix weighting the intended parameter updates, LWF computes forgetting confidence to evaluate self-generated knowledge regarding the forgetting task, and consequently, knowledge with high confidence is periodically unlearned during fine-tuning. Our experiments demonstrate that, although thoroughly uncovering the mechanisms of knowledge interaction remains challenging in pre-trained language models, applying graceful forgetting can contribute to enhanced fine-tuning performance. 
\end{abstract}

\section{Introduction}
\label{sec:intro}
In recent years, the \textit{pretrain-finetune} paradigm has emerged as a dominant framework across natural language processing (NLP) tasks and various other domains~\citep{pretrain-survey}. This approach involves pre-training the model on large-scale corpora and subsequently fine-tuning it on smaller, task-specific datasets to adapt to downstream applications. Its effectiveness has been evidenced by the success of prominent pre-trained models such as BERT~\citep{bert}, GPT~\citep{GPT3}, and T5~\citep{T5}. And these models have become the backbone of many state-of-the-art AI applications~\citep{gpt4, stable-diffuusion}.

Despite offering compelling benefits such as data efficiency and reusability, this well-established paradigm continues to face a long-standing and prevalent issue, \textit{negative transfer}, which surfaces in a new guise. While the typical interpretation of negative transfer in transfer learning refers to the performance degradation when learning conflicting tasks simultaneously or sequentially~\citep{NT-survey}, its manifestation in the context of the pretrain-finetune paradigm takes on a different form, \textit{the negative contribution of some pre-trained knowledge to the target fine-tuning task}~\citep{negative-transfer}. This problem highlights a critical limitation of vanilla fine-tuning: treating all pre-trained knowledge indiscriminately is not always the optimal practice. 

To address this issue, a promising approach is to suppress the influence of potentially harmful knowledge, a strategy known as \textit{graceful forgetting} (or active forgetting). Originating in neuroscience, this concept describes a memory mechanism in biological intelligence where the ability to acquire new knowledge is enhanced by selective elimination of irrelevant or outdated information~\citep{bio-active-forget}. Recent advances have demonstrated the feasibility of emulating this mechanism in machine learning models~\citep{ML-forget1, ML-forget2}, leading to its adoption in various studies aimed at enhancing learning plasticity~\citep{AFEC, multi-lingual-forget, adaptive-plasticity, SRS}. 


However, most existing graceful forgetting methods are either tailored to vision tasks or designed for non-autoregressive models, making their efforts incompatible or less effective when migrating to generative language models. This discrepancy primarily stems from the ambiguous knowledge boundaries inherent in language generation, which significantly complicate the identification of explicit and granular inter-task correlations~\citep{intermedia-tasks}.

To address this gap, in this paper we investigate the graceful forgetting in generative language models. The central question guiding our study is: \textit{can generative language models achieve more effective fine-tuning by gracefully forgetting unnecessary knowledge?} To answer this question, we propose a framework called Learning with Forgetting (\textbf{LWF}) to enable graceful forgetting in generative language models. Beginning with addressing the inaccessibility of pre-trained data, LWF leverages the inherent capabilities of generative models by expressing knowledge related to the forgetting task through self-generated texts. Furthermore, given the difficulty of identifying task-level correlations, LWF computes a data-wise \textit{forgetting confidence} for each input by weighting the intended parameter updates with the Fisher Information Matrix. Based on this metric, LWF selects high-confidence data points and integrates machine unlearning techniques to periodically remove associated knowledge during the fine-tuning process. 

To the best of our knowledge, LWF is the first systematic exploration to enhance the learning plasticity of pretrained generative language models by graceful forgetting. Through extensive experiments and analyses, we demonstrate the feasibility of improving fine-tuning performance through graceful forgetting. Hopefully, our empirical findings will contribute to a better understanding of this emerging topic and offer inspiration for future investigation and innovation.

\section{Related Work}
\label{sec:literature}
In this section, we provide an overview of how the critical concepts in our work are interpreted and applied across a broader research landscape, and analyze their commonalities and distinctions.

\subsection{Negative Transfer}
\label{relate:NT}
Negative transfer is a prevalent issue across multiple fields, with its interpretation and definition often varying depending on the context.

In Multi-Task Learning (MTL), negative transfer refers to the performance degradation caused by mutual interference among conflicting tasks~\citep{NT-diffusion}. While the primary objective of MTL is to learn multiple tasks simultaneously, methods aimed at mitigating negative transfer typically focus on quantifying inter-task relationships using sophisticated metrics like gradient directions~\citep{fork-merge, learning2learn} or signal-to-noise ratio~\citep{NT-diffusion}. Based on these metrics, tasks can be clustered into separate groups~\citep{model-zoo} to reduce learning conflicts. In the MTL context, strategies that involve sacrificing part of the model’s capacity, such as forgetting, are typically not regarded as appropriate solutions. 

Another related domain is Continual Learning (CL), where different tasks are learned sequentially. In CL, negative transfer is bi-directional: it can refer to the interference of previously acquired knowledge with the learning of new tasks, or to the forgetting of past knowledge caused by learning new ones~\citep{learning-curves}. While most CL methods focus on maintaining memory stability when learning new tasks~\citep{EWC, progress-compress}, recent studies have highlighted that this stability often comes at the cost of reduced learning plasticity. In response, these approaches actively weaken the preservation strength of past memory, seeking a balance between memory stability and learning plasticity~\citep{adaptive-plasticity, AFEC, progress-compress}, which can be seen as an implicit form of forgetting. 

In our context, the pretrain-finetune paradigm, negative transfer denotes the detrimental influence of certain pre-trained knowledge on the target fine-tuning task. A distinguishing characteristic of negative transfer in this paradigm is that the pre-training data is typically inaccessible during fine-tuning, which renders many countermeasures used in MTL and CL ineffective or inapplicable. 

\subsection{Graceful Forgetting}
In many domains, forgetting is traditionally viewed as an undesirable phenomenon, reflecting a failure to retain previously acquired knowledge. This concern is particularly prominent in CL, where one of the primary objectives is to overcome catastrophic forgetting~\citep{EWC}.  

However, recent studies have argued that striving for an omniscient model may be impractical due to limited model capacity and inevitable knowledge conflicts~\citep{ML-forget1, ML-forget2}. Drawing inspiration from neuroscience~\citep{bio-active-forget}, an increasing number of studies have explored the potential of improving learning plasticity through actively forgetting irrelevant or outdated knowledge. In the context of CL, \citet{AFEC} proposed a synaptic expansion-convergence mechanism to selectively forget preserved knowledge. \citet{adaptive-plasticity} realized controllable learning plasticity through gradient projection. In the context of fine-tuning, \citet{BSS} and \citet{SRS} incorporate model structural shrinkage regulation to enable implicit forgetting. While most of these methods were initially designed for non-autoregressive tasks, \citet{multi-lingual-forget} extended graceful forgetting to the LLM pre-training stage, enhancing multi-lingual ability by refreshing models. \citet{F-learning} proposed a forgetting-before-learning method to achieve knowledge edition for LLMs.

Despite these efforts, existing graceful forgetting approaches are either inapplicable or significantly less effective when adapted to promoting fine-tuning of generative language models.

\subsection{Machine Unlearning}
Machine unlearning has emerged as a vibrant and rapidly evolving research area focused on selectively removing specific data, patterns, or knowledge from trained models~\citep{unlearn-survey}. In the context of generative language models, unlearning is frequently employed to align model behavior with human values—such as safeguarding user privacy~\citep{user-protect}, eradicating harmful or biased content~\citep{unlearn-survey2}, and mitigating hallucinations~\citep{unlearn-survey}. A range of unlearning strategies have been proposed for this purpose, including gradient ascent~\citep{unlearning-ga}, localization-informed unlearning~\citep{unlearning-ga}, and influence function-based approaches~\citep{unlearning-influence}, among others. While current unlearning research about generative language models primarily focuses on eradicating undesirable behaviors, our work repurposes unlearning as a mechanism to achieve graceful forgetting, thereby enhancing the plasticity of fine-tuning. In essence, we leverage unlearning for better learning.
\section{Methodology}
\begin{figure*}[t]
    \centering
    \includegraphics[width=\textwidth]{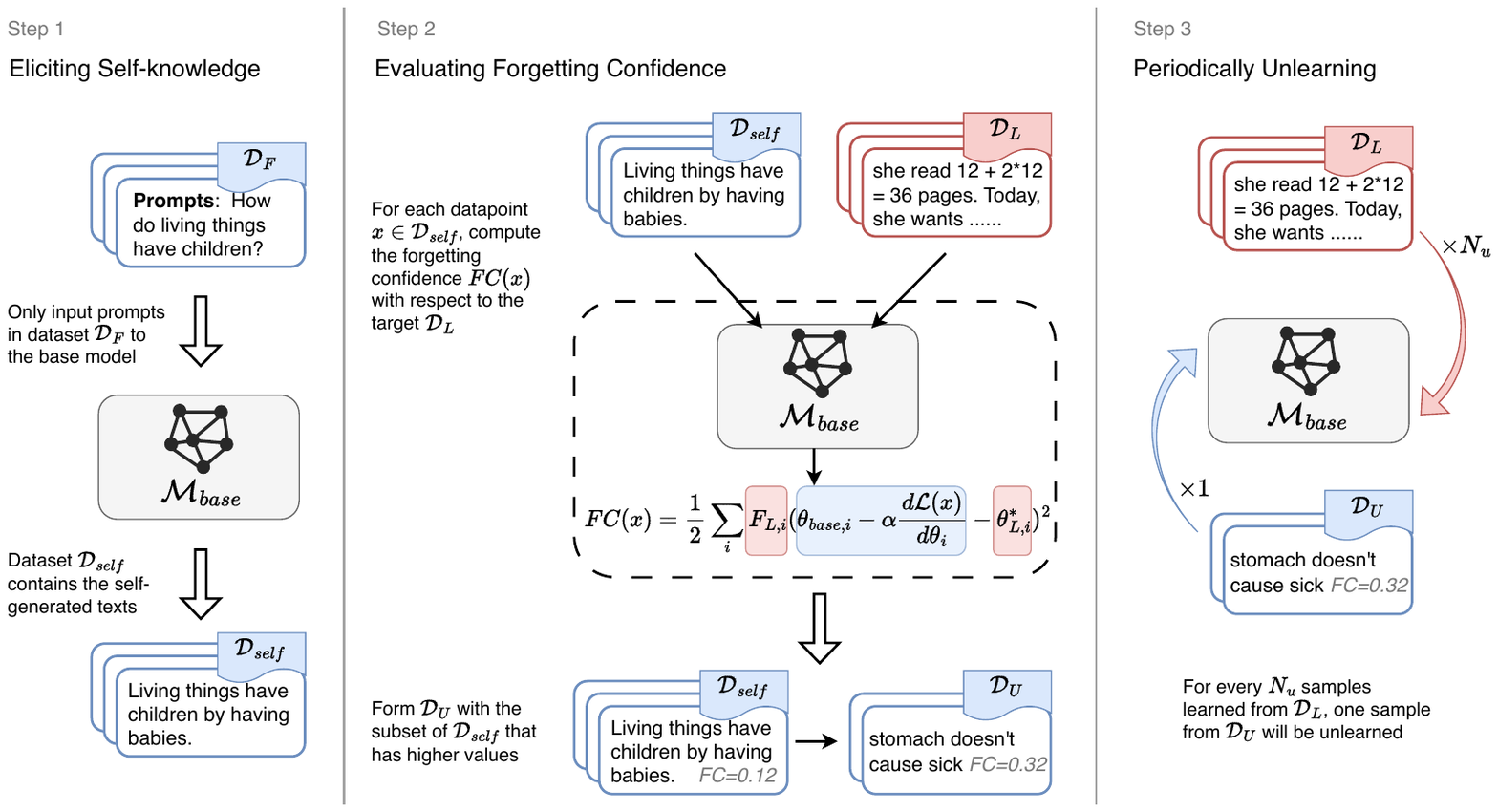}
    \caption{The overview of the LWF framework. Given the forgetting task $\mathcal{D}_F$ and learning task $\mathcal{D}_L$, LWF first constructs $\mathcal{D}_{self}$ through self-generated texts to represent the knowledge regarding the forgetting task. Then, with the Fisher Information Matrix $F_L$ and the optimal parameters of the learning task approximated from $\mathcal{D}_L$, LWF calculates forgetting confidence for each data point in $\mathcal{D}_{self}$. Finally, data points with high forgetting confidence are selected for unlearning, represented by $\mathcal{D}_U$. The unlearning process is integrated into the fine-tuning process of $\mathcal{D}_L$ and is executed periodically at intervals of $N_u$.}
    \label{fig:overview}
\end{figure*}
In this section, we detail the implementation of our framework for graceful forgetting in generative language models, Learning With Forgetting (\textbf{LWF}). It consists of three components: eliciting self-knowledge, evaluating forgetting confidence, and periodically unlearning. Fig~\ref{fig:overview} illustrates the overview. For the sake of convenience in exposition, we use $\mathcal{D}_L$ to represent the learning task and $\mathcal{D}_F$ to denote the forgetting task. It is important to note, however, that the framework is task quantity-agnostic, which will be elaborated in Section~\ref{sec:exp}.

\subsection{Eliciting Self-Knowledge}
The first step in forgetting specific knowledge is to acquire its representation. However, as discussed in Sec~\ref{relate:NT}, the pre-training corpus is typically inaccessible in practice, making it uncertain whether $\mathcal{D}_F$ can adequately represent the model's knowledge. To address this issue, we exploit the inherent characteristics of generative language models as an alternative: leveraging self-generated data. Specifically, we input the prompts (e.g., questions or instructions) from $\mathcal{D}_F$ into the base model $\mathcal{M}_{base}$ and collect its responses to form the unlearning dataset, which we donate as $\mathcal{D}_{self}$. This approach also enables LWF to utilize unlabeled datasets.

\subsection{Evaluating Forgetting Confidence}
\label{sec:fc}
Not all kinds of knowledge oblivion contribute positively to model adaptation. To ensure that forgetting improves, rather than degrades, the fine-tuning performance, we propose a confidence metric to evaluate the safety of unlearning specific knowledge. Considering the semantic richness of natural language, a task-level metric proves insufficiently nuanced~\citep{intermedia-tasks}. Therefore, we define the \textit{forgetting confidence} at the individual data point level, enabling fine-grained assessment of what should be forgotten.

For a generated text $x$ in $\mathcal{D}_{self}$, the posteriori $P(\mathcal{D}_L|x)$ intuitively reflects to what extent $\mathcal{D}_L$ and $x$ are synergistic. The lower $P(\mathcal{D}_L|x)$ is, the more likely $x$ is conflicted with $\mathcal{D}_{self}$. Considering $P(\mathcal{D}_L|x)$ is computationally intractable, we use $P(\mathcal{D}_L|\theta^*(x))$ as a surrogate, where 
\begin{equation}
    \label{equ:forget-theta}
    \theta^*(x) = \mathop{\arg\max}\limits_{\theta} P(\theta|x)
\end{equation}
Here $\theta$ represents the model parameters. Since only the relative value is required, we can use $P(\theta^*(x)|\mathcal{D}_L)$ to represent $P(\mathcal{D}_L|\theta^*(x))$, as the two are positively proportional according to the Bayes' Theorem. Based on this, we define the forgetting confidence as:
\begin{equation}
    \label{equ:fc}
    FC(x) \propto - \log P(\theta^*(x)|\mathcal{D}_L)
\end{equation}

Following prior works~\citep{EWC,AFEC}, we assume $P(\theta|\mathcal{D}_L)$ as a Gaussian distribution centered at $\theta^*_L = \mathop{\arg\max}\limits_{\theta} P(\theta|\mathcal{D_L})$, and this distribution can be approximated using a second-order Taylor expansion around $\theta^*_L$:
\begin{equation}
    \label{equ:tylor}
    \begin{aligned}
        \log P(\theta|\mathcal{D}_L) &\approx \frac{1}{2}(\theta - \theta^*_L)^T\\&(\frac{\partial^2\log P(\theta|\mathcal{D}_L)}{\partial^2\theta}|_{\theta^*_L})(\theta - \theta^*_L)\\
    \end{aligned}
\end{equation}
In practice, we integrate Equation~\ref{equ:fc} with Equation~\ref{equ:tylor} and use a single-step update from the base model to represent $\theta^*(x)$, thereby reducing computational costs:
\begin{equation}
    \label{equ:FC-practical}
    FC(x) = \frac{1}{2}\sum_i F_{L,i}(\theta_{base,i}-\alpha\frac{d \mathcal{L}(x)}{d\theta_i}-\theta^*_{L,i})^2
\end{equation}

$F_L$ represents the Fisher Information Matrix (FIM), which is the negative expectation of the Hessian Matrix in Equation~\ref{equ:tylor}. The parameters of the base model are represented by $\theta_{base}$, and $\mathcal{L}(x)$ refers to the cross-entropy loss of $x$. $\alpha$ controls the margin of the single-step update. $\theta^*_L$ is obtained by training the base model on $\mathcal{D}_L$. We include a detailed step-by-step deduction in Appendix~\ref{sec:deduction} and conduct a sensitivity analysis about the approximation error of the one-step update in Appendix~\ref{sec:approx}.

Intuitively, Equation~\ref{equ:FC-practical} measures the conflict between $x$ and $\mathcal{D}_L$ by evaluating the alignment between the intended parameter update induced by $x$ and the target $\theta^*_L$. The FIM $F_L$ serves as a weighting mechanism that captures the relative importance of each parameter.

\subsection{Periodically Unlearning} 
Due to the well-documented instability of machine unlearning~\citep{unlearn-survey,unlearn-survey2}, directly unlearning samples from $\mathcal{D}_{self}$ may yield inconsistent performance gains, particularly because the selected samples are only potentially conflicting with the target task, not definitively so. To mitigate the instability introduced by unlearning, we adopt a \textit{"periodically unlearning"} strategy that interleaves learning and unlearning simultaneously throughout one training process.

Specifically, we introduce a fixed unlearning interval $N_u$, such that the unlearning is applied every $N_u$ learning steps. For instance, when $N_u=7$, we first select a subset $\mathcal{D}_U$ from $\mathcal{D}_{self}$, consisting of the top $\frac{|\mathcal{D}_L|}{7}$ samples of the highest forgetting confidence scores $FC(x)$. During training, both of the learning dataset $\mathcal{D}_L$ and the unlearning dataset $\mathcal{D}_U$ are utilized: for every $7$ learning samples drawn from $\mathcal{D}_L$, one sample from $\mathcal{D}_U$ is unlearned. This balanced interleaving helps prevent the forgetting process from impairing the learning dynamics.

We use \textit{Gradient Ascent}~\citep{unlearning} as our unlearning algorithm, which merely involves negating the loss function. Specifically, for a periodic batch $\mathcal{X}=\{x_1^l,\dots,x_{N_u}^l,x^u\}$ where $\{x_1^l,\dots,x_{N_u}^l\}\subset \mathcal{D}_L$ and $x^u\in\mathcal{D}_U$, the loss can be written as:
\begin{equation}
\label{equ:total-loss}
    \mathcal{L}_{pu}(\mathcal{X}) = \sum_{x\in\{x_1^l,\dots,x_{N_u}^l\}} \mathcal{L}(x) - \beta\mathcal{L}(x^u)
\end{equation}
where $\mathcal{L}$ is the \textit{sft} loss and $\beta$ is the unlearning rate.
\section{Experiments}
\label{sec:exp}
\subsection{Setup}
\subsubsection{Datasets}
To evaluate the effectiveness of LWF, we apply our method to domain-specific question-answering tasks. This choice is motivated by the well-established evaluation metrics (i.e., accuracy) and the relatively clear delineation of knowledge boundaries across domains. We further discuss task generalizability in Appendix~\ref{task-generalizability}. 

We select five datasets spanning diverse domains: \textbf{gsm8k}~\citep{gsm8k} for mathematical reasoning; \textbf{qasc}~\citep{qasc} for elementary science; \textbf{sst5}~\citep{sst5} for sentiment classification; \textbf{dental}, a subset of MedMCQA~\citep{medmcqa} focusing on dental knowledge; and \textbf{psychol}, a subset of MMLU~\citep{mmlu} covering psychology.

\subsubsection{Baselines}
As discussed in Sections~\ref{sec:intro} and~\ref{sec:literature}, although research on forgetting in broader contexts is not lacking, few existing graceful forgetting methods are specifically designed to enhance fine-tuning plasticity for pre-trained generative language models, which makes effective method migration challenging. To place LWF in a broader methodological context, we select three baseline approaches originally developed for different settings but adaptable to ours with minimal modifications: BSS~\citep{BSS} and SRS~\citep{SRS}, two structural regularization methods that mitigate negative transfer from pretrained knowledge in non-autoregressive models; and F-learning~\citep{F-learning}, an active unlearning technique proposed for model editing in large language models.

\subsubsection{Implementation Details}
We use Llama3.2-1B~\citep{Llama} as the base model. For eliciting self-knowledge, we employ a 3-shot prompt concatenated with the input question, use greedy decoding, and limit the maximum number of generated tokens to 256. For computing the forgetting confidence, we set the one-step update coefficient $\alpha = 1 \times 10^{-2}$ (as defined in Equation~\ref{equ:FC-practical}). To maintain gradient coherence during periodic unlearning, we interleave the datasets $\mathcal{D}_L$ and $\mathcal{D}_U$ by including one sample from $\mathcal{D}_U$ for every $N_u$ samples from $\mathcal{D}_L$. We set $N_u=7$ and choose $\beta$ (from Equation~\ref{equ:total-loss}) as either 0.1 or 0.05, depending on the forgetting task. The training batch size is 4, which results in one unlearning sample appearing every two batches. We optimize using AdamW~\citep{AdamW} with a learning rate of $1 \times 10^{-5}$, and train for one full epoch. All experiments are conducted on eight NVIDIA RTX 4090 GPUs with full-parameter fine-tuning. For further implementation details, please refer to Appendix~\ref{sec:detail} and our source code repository \url{https://github.com/rubickkcibur/LWF}. 
 
\subsection{Results on Question Answering}
\begin{table}
    \resizebox{\columnwidth}{!}{
    \begin{tabular}{cccccc}
    \toprule
     & gsm8k & qasc & sst5 & dental & psychol \\
    \midrule
    none & 19.71 & 42.98 & 49.55 & 36.87 & 46.42 \\
    \midrule
    gsm8k & - & +4.03\% & +2.83\% & +1.46\% & +6.33\% \\
    qasc & +5.38\% & - & +2.54\% & -4.53\% & +5.54\% \\
    sst5 & +2.67\% & +3.02\% & - & +0.22\% & +0.41\% \\
    dental & +10.40\% & +5.28\% & +2.10\% & - & +1.59\% \\
    psychol & +1.17\% & +2.00\% & +1.27\% & -4.10\% & - \\
    \midrule
    mixed & +6.95\% & +5.54\% & +2.10\% & +1.46\% & +7.93\% \\
    \bottomrule
    \end{tabular}
    }
    \caption{Results on domain-specific question answering. Each column shares the same learning task and the rows represent different forgetting tasks. All percentages are calculated based on \textit{none}.}
    \label{tab:qa-result}
\end{table}
\begin{table}
    \resizebox{\columnwidth}{!}{
    \begin{tabular}{cccccc}
    \toprule
     & gsm8k & qasc & sst5 & dental & psychol \\
    \midrule
    none & 56.56 & 68.36 & 55.52 & 49.39 & 80.00 \\
    \midrule
    gsm8k & - & +5.37\% & +0.97\% & +0.78\% & +2.06\% \\
    qasc & +1.61\% & - & +0.74\% & -3.06\% & +0.46\% \\
    sst5 & +7.37\% & +2.68\% & - & +0.00\% & +2.75\% \\
    dental & +1.20\% & +1.26\% & +1.30\% & - & +0.91\% \\
    psychol & +1.47\% & +4.42\% & +0.24\% & -0.14\% & - \\
    \midrule
    mixed & +4.56\% & +7.90\% & +1.06\% & +0.78\% & +1.60\% \\
    \bottomrule
    \end{tabular}
    }
    \caption{Results on domain-specific question answering with a larger language model.}
    \label{tab:qa-result-8B}
\end{table}
\begin{table}
    \resizebox{\columnwidth}{!}{
    \begin{tabular}{ccccccc}
    \toprule
     & gsm8k & qasc & sst5 & dental & psychol & AVG.\\
    \midrule
    vanilla-FT & 19.71 & 42.98 & 49.55 & 36.87 & 46.42 & 39.12\\
    \midrule
    BSS & 20.39 & 44.28 & 49.73 & 35.51 & 44.77 & 38.94\\
    SRS & 17.36 & 40.28 & 50.50 & 35.05 & 46.61 & 37.96\\
    F-learning & 17.29 & 46.44 & 51.31 & 34.07 & 42.75 & 38.37\\
    \midrule
    LWF-mixed & 21.08 & 45.36 & 50.59 & 37.41 & 50.10 & 40.91\\
    \bottomrule
    \end{tabular}
    }
    \caption{Accuracy results of three adapted baseline methods compared to vanilla fine-tuning and LWF under \textit{mixed} setting.}
    \label{tab:baselines}
\end{table}
Table~\ref{tab:qa-result} presents the results on question-answering tasks. Each column corresponds to a distinct learning task, and each row represents a specific forgetting scenario. For example, the value $+5.38\%$ at the intersection of the gsm8k column and the qasc row indicates that, when fine-tuning on gsm8k, unlearning qasc using LWF improves performance by $5.38\%$ compared to vanilla fine-tuning. In particular, the first row, labeled \textit{none}, reports the vanilla fine-tuning results (\textit{i.e.}, no unlearning). The last row, labeled \textit{mixed}, represents a forgetting scenario in which all datasets except the target learning dataset are jointly unlearned. All entries except those in the \textit{none} row report the percentage improvement in accuracy relative to the corresponding vanilla fine-tuning baseline. 

As shown in the results, in most cases, LWF improves performance on the learning task compared to vanilla fine-tuning. Exceptions occur when learning dental while forgetting qasc or psychol, which we believe is attributed to the low forgetting confidence of self-generated samples. This is further evidenced by the results of the \textit{mixed}, where consistent improvements across all learning tasks suggest that combining diverse forgetting datasets increases the pool of high-confidence candidates, thereby enhancing the likelihood of performance gains. \textit{The mixed setting is also recommended in practice for stable effectiveness}.

Table~\ref{tab:baselines} compares LWF-mixed with the three adapted baseline methods. As shown, although one of the baselines occasionally achieves the highest improvement in certain cases, on average, all three perform worse than both LWF-mixed and even vanilla fine-tuning. We attribute this to the fundamental mismatch between these methods and our setting. Specifically, BSS and SRS were designed for non-autoregressive models and are not optimized for autoregressive generation tasks. Meanwhile, F-learning targets model editing scenarios in which the knowledge to be unlearned is explicitly known a priori to be detrimental or outdated, a condition that does not hold in our context, where forgetting targets are not necessarily harmful.


We also examine the side effects of LWF, \textit{i.e.}, its impact on datasets that are neither part of the learning task nor the forgetting task. Details are discussed in Appendix~\ref{sec:side-effect}.

\subsection{Scalability Analysis}
A natural question is: \textit{Would LWF become less effective in larger generative language models, given that their increased parameter capacity may better accommodate conflicting knowledge?} To investigate this, we apply LWF to the Llama3-8B model. As shown in Table~\ref{tab:qa-result-8B}, LWF can still improve fine-tuning performance in most cases. While the magnitude of relative improvement has declined overall, this trend is partly due to the stronger baseline performance of the larger model. Notably, the discipline gained from smaller model settings generalizes well to the larger: the \textit{mixed} setting is still the best choice to achieve stable effectiveness.

\subsection{Analysis on Forgetting Confidence}
\begin{figure}[t]
    \centering
    \includegraphics[width=\columnwidth]{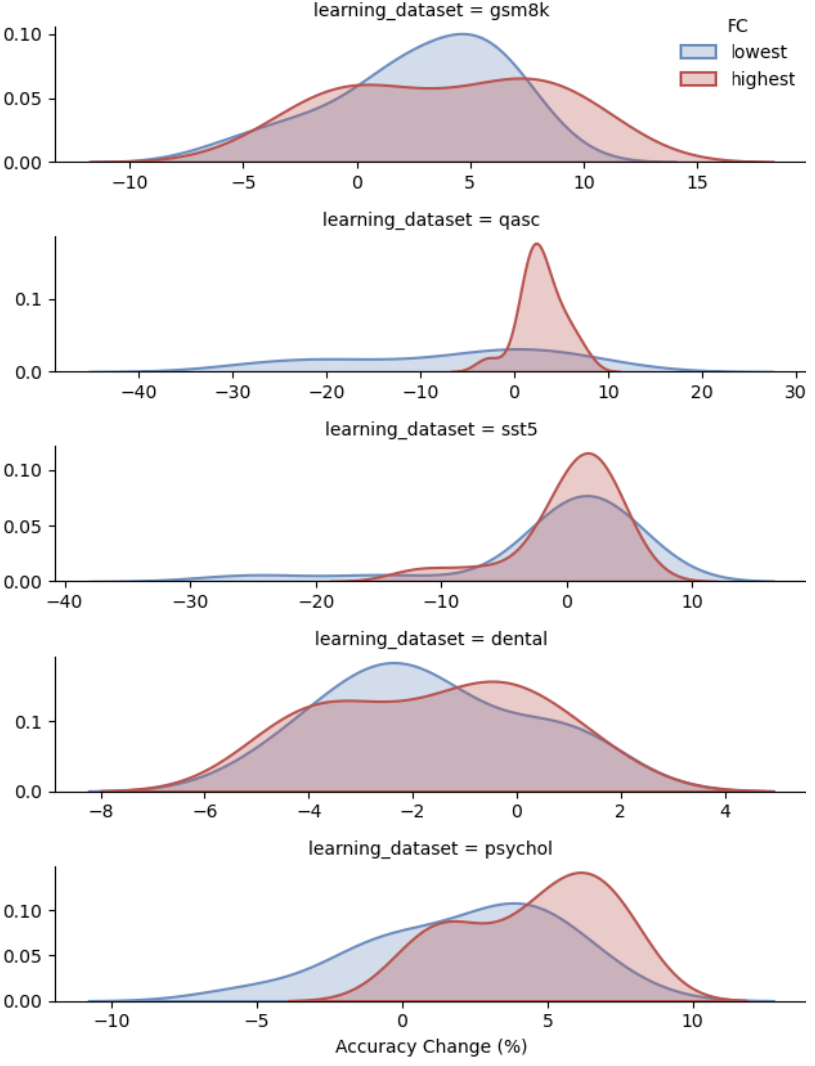}
    \caption{Distribution of accuracy changes between two filtering strategies. The $y$-axis represents distribution density. All percentages are calculated based on vanilla fine-tuning.}
    \label{fig:violin}
\end{figure}
A central component of LWF is the computation of forgetting confidence, which estimates the likelihood that forgetting a specific data point will benefit the learning task. While we propose a computable metric $FC(x)$ (in Sec~\ref{sec:fc}), it involves approximations and does not provide an exact mapping between gradient updates and performance outcomes. As a result, $FC(x)$ should be viewed as a heuristic rather than an absolute measure. In practice, we also observe that unlearning data with relatively low $FC$ does not necessarily lead to performance degradation.

To gain deeper statistical insight into the role of forgetting confidence, we design an ablation study using an inverse filtering strategy: selecting data with the lowest forgetting confidence. For each learning task, we compute the percentage change in accuracy relative to vanilla fine-tuning, across all forgetting tasks. To increase the sample size, we aggregate results over four unlearning rates: $\beta \in \{0.05, 0.10, 0.20, 0.25\}$.

Fig~\ref{fig:violin} shows the distribution of accuracy changes under the two filtering strategies. The \textit{red} region corresponds to unlearning data with the \textit{highest} $FC$, while the \textit{blue} region represents unlearning data with the lowest. The $x$-axis indicates the percentage change in accuracy relative to vanilla fine-tuning. As shown, unlearning high-$FC$ data generally outperforms the alternative in two aspects: higher average performance gain and greater stability. Specifically, high-$FC$ unlearning yields smaller variance and a narrower overall range, whereas unlearning low-$FC$ data leads to more volatile outcomes, including severe performance drops in extreme cases. In summary, prioritizing data with high forgetting confidence results in consistently better and more robust performance, making it a more reliable strategy in practice.

\subsection{Ablation on Periodically unlearning}
\begin{table}
    \resizebox{\columnwidth}{!}{
    \begin{tabular}{cccccc}
    \toprule
    $\mathcal{D}_F$ & gsm8k & qasc & sst5 & dental & psychol \\
    \midrule
    none & 19.71 & 42.98 & 49.55 & 36.87 & 46.42 \\
    \midrule
    gsm8k & - & -65.6\% & +0.5\% & -0.8\% & -9.9\% \\
    qasc & +5.0\% & - & +0.4\% & -7.4\% & +3.6\% \\
    sst5 & +4.3\% & +7.8\% & - & -3.5\% & +9.1\% \\
    dental & -8.5\% & -26.6\% & +1.2\% & - & +0.0\% \\
    psychol & -6.5\% & -8.3\% & -21.7\% & -4.9\% & - \\
    \midrule
    mixed & -3.5\% & -73.9\% & +2.3\% & +0.8\% & -7.9\% \\
    \bottomrule
    \end{tabular}
    }
    \caption{Results of \textit{ahead unlearning}, where unlearning is completed before fine-tuning, as an ablation study for periodically unlearning.}
    \label{tab:ahead}
\end{table}
\begin{table}
    \resizebox{\columnwidth}{!}{
    \begin{tabular}{cccccc}
    \toprule
    $\mathcal{D}_F$ & gsm8k & qasc & sst5 & dental & psychol \\
    \midrule
    none & 19.71 & 42.98 & 49.55 & 36.87 & 46.42 \\
    \midrule
    gsm8k & - & -12.6\% & -0.1\% & -4.9\% & +10.3\% \\
    qasc & -8.8\% & - & +3.9\% & -2.0\% & +7.7\% \\
    sst5 & -8.8\% & -10.8\% & - & -4.3\% & +3.6\% \\
    dental & -6.5\% & -6.9\% & +1.8\% & - & +7.5\% \\
    psychol & -5.0\% & -13.1\% & +2.1\% & -2.7\% & - \\
    \midrule
    mixed & -6.5\% & -10.1\% & -0.5\% & -4.9\% & +9.5\% \\
    \bottomrule
    \end{tabular}
    }
    \caption{Results of \textit{randomly unlearning}, where unlearning is randomly executed during fine-tuning, as an ablation study for periodically unlearning.}
    \label{tab:random}
\end{table}
To alleviate the vulnerability of machine unlearning, we propose the periodically unlearning strategy to stabilize the training process. In this section, we conduct an ablation study to demonstrate that periodically unlearning is the most effective strategy for integrating learning and unlearning.

We compare two alternative unlearning strategies. The first performs unlearning entirely before the learning process, which we refer to as \textit{ahead unlearning}. The second interleaves unlearning steps randomly during training, termed \textit{randomly unlearning}. For fairness, all three strategies, including \textit{periodically unlearning}, maintain the same ratio of learning to unlearning samples.

Table~\ref{tab:ahead} and Table~\ref{tab:random} present the results of \textit{ahead unlearning} and \textit{randomly unlearning} respectively. As we can see, both strategies are generally much less effective than \textit{periodically unlearning} (Table~\ref{tab:qa-result}), with most learning-forgetting combinations resulting in performance degradation. Notably, \textit{ahead unlearning} exhibits several extremely detrimental cases, primarily attributed to the unintended disruption to foundational pre-trained knowledge caused by premature unlearning steps. If such critical knowledge is compromised, the subsequent fine-tuning process may suffer severe performance loss. 
In conclusion, the interleaved learning and unlearning combination is better than conducting them separately, and switching them periodically is superior to randomly.

\subsection{Analysis on the Forgotten Task}
\begin{figure}[t]
    \centering
    \includegraphics[width=0.95\columnwidth]{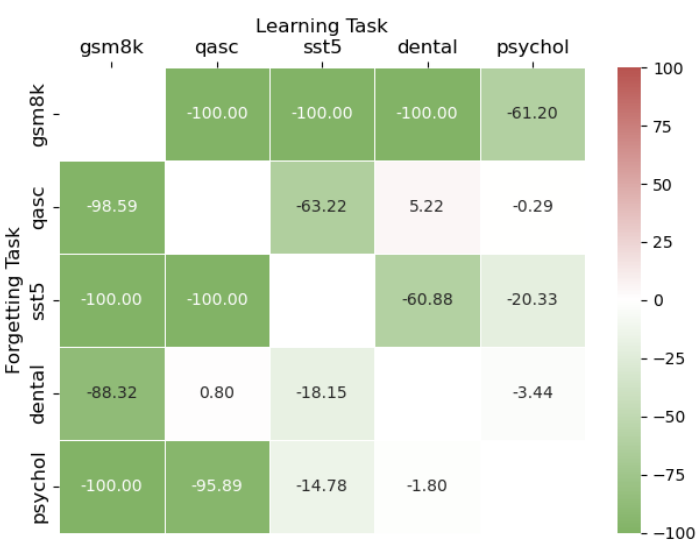}
    \caption{Accuracy change percentage of the forgetting task across different learning-forgetting combinations. Percentages are computed based on vanilla fine-tuning.}
    \label{fig:acc-drop}
\end{figure}
\begin{figure}[t]
    \centering
    \includegraphics[width=0.95\columnwidth]{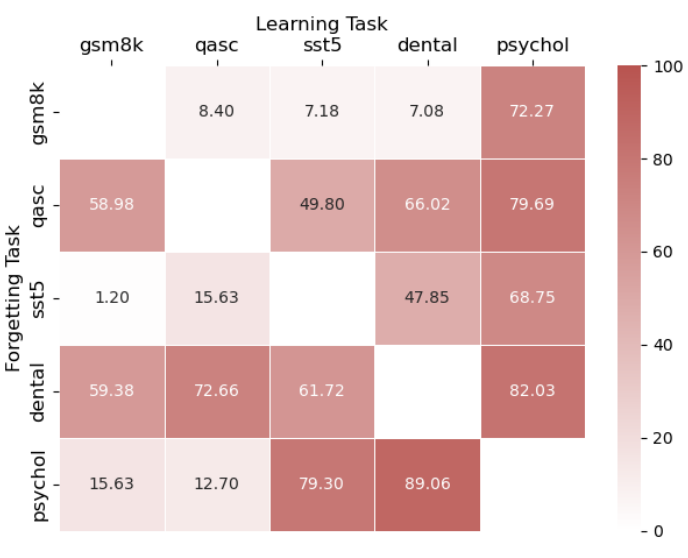}
    \caption{Cosine similarity between the outputs of forgetting tasks generated by the vanilla fine-tuned model and LWF resulting model. Values are multiplied by 100.}
    \label{fig:sim-drop}
\end{figure}
\begin{figure}[t]
    \centering
    \includegraphics[width=0.95\columnwidth]{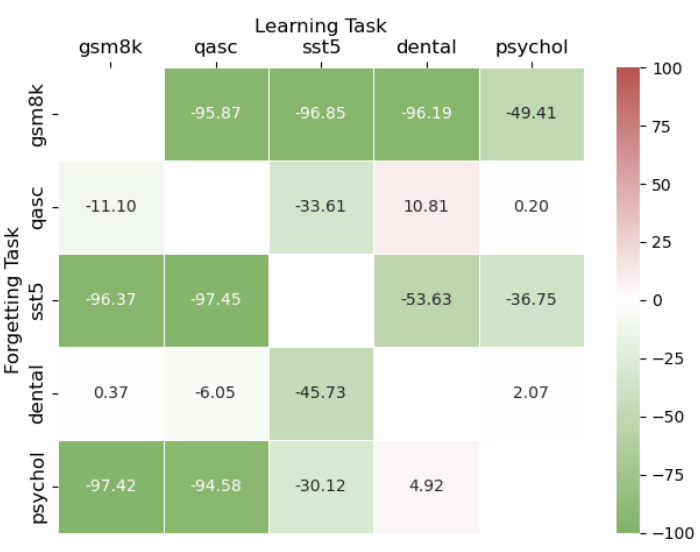}
    \caption{TTR change percentage of the forgetting task across different learning-forgetting combinations. Percentages are computed based on vanilla fine-tuning.}
    \label{fig:ttr-change}
\end{figure}
In this section, we examine how model performance evolves on the forgetting task before and after applying LWF. While it may seem intuitive that the model’s knowledge of the forgetting task would be substantially erased, the empirical results reveal a more nuanced picture. Figure~\ref{fig:acc-drop} shows a heatmap of accuracy changes (in percentage points) on the forgetting tasks relative to vanilla fine-tuning. For instance, when the learning task is psychol and the forgetting task is gsm8k, the value $-61.20$ indicates a $61.2\%$ drop in performance on gsm8k. As the figure illustrates, although accuracy generally declines across most learning-forgetting pairs, the extent of degradation varies significantly. In approximately half of the cases, performance drops by nearly $100\%$, suggesting near-complete unlearning; in others, the decline is much smaller, particularly when either the learning or the forgetting task involves dental or psychol. 

Additionally, we noticed that the accuracy alone merely reflects whether the final answer is correct, which is insufficient to fully capture the nuanced changes in the generated outputs of generative models. To gain deeper insights, we further analyze the semantic shifts in the responses. Specifically, we use SimCSE model~\citep{simcse} to obtain the sentence vectors of the responses generated by the vanilla fine-tuned model and LWF model when answering the same forgetting task questions.We then compute the cosine similarity between the resulting sentence embeddings to quantify the degree of semantic change. Results are presented in Fig~\ref{fig:sim-drop}. 

Given that a cosine similarity score above $80\%$ is typically required to confidently assert semantic similarity between two sentences, the results indicate that most learning-forgetting combinations exhibit substantial semantic changes. Notably, the cases where similarity approaches or exceeds this threshold largely coincide with those showing minimal accuracy drops, \textit{i.e.}, combinations involving the dental or psychol. We believe this phenomenon may be attributed to that dental and psychol are inherently more complex than the other tasks. Acquiring or forgetting these domains likely requires engagement with richer, more structured knowledge, which in turn makes them more resistant to severe forgetting.

We also evaluate changes in lexical diversity, as shown in Fig~\ref{fig:ttr-change}, where Type Token Ratio (TTR) is the metric. Similar to the trends observed in accuracy changes, the TTR experiences a significant decline in most combinations, and the cases maintaining high accuracy and semantic similarity also largely preserve their lexical diversity.

\subsection{Multi-Task Learning}
\begin{figure}[t]
    \centering
    \includegraphics[width=\columnwidth]{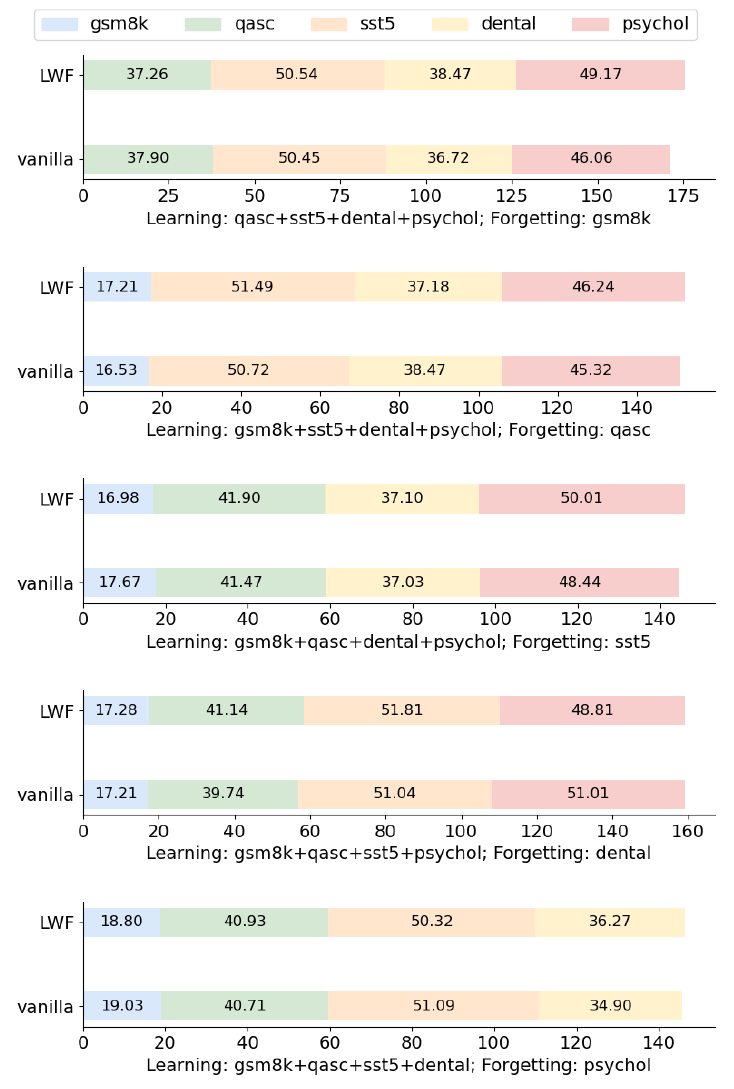}
    \caption{Accuracy results in the multi-task learning setting. Labeled below each subplot are the forgetting task and learning tasks.}
    \label{fig:multi}
\end{figure}
In this section, we examine the performance of LWF in multi-task learning scenarios. Specifically, we treat one of the five datasets as the forgetting task and use the remaining four as learning tasks. To mitigate catastrophic forgetting of earlier tasks, we train on a balanced mixture of all learning datasets. Fig~\ref{fig:multi} compares the overall multi-task accuracy between LWF and vanilla fine-tuning. As we can see, while not all individual learning task gets improved, LWF outperforms vanilla fine-tuning in general.

These results highlight the complexity of multi-task learning. Although LWF enables overall performance gains through controlled forgetting, the interactions among learning tasks are non-trivial. Improvements in some tasks may come at the expense of others.


\section{Conclusion}
In this paper, we propose a novel framework, Learning With Forgetting (LWF), to achieve graceful forgetting in generative language models. LWF addresses the inaccessibility of pre-trained data by leveraging self-generated knowledge, calculates forgetting confidence for each data point by weighting the intended parameter update with the Fisher Information Matrix, and employs gradient ascent to periodically unlearn high-confidence data during fine-tuning. Empirical results on domain-specific question-answering tasks demonstrate the effectiveness of LWF. Furthermore, we conduct extensive experiments to analyze the contribution of each component of LWF, the effects of forgetting specific tasks, and the framework's performance in learning or forgetting multiple tasks. While fully elucidating the mechanisms of inter-task interactions and achieving optimal graceful forgetting still need great effort, we hope our work provides valuable insights into this emerging area and inspires future research endeavors.

\section*{Limitation}
LWF still has several limitations that warrant further investigation. First, our proposed forgetting confidence metric is based on analyzing intended parameter updates to select data for unlearning. While empirical results demonstrate its statistical effectiveness, performance may degrade when the pool of candidate forgetting samples is small. Since quantifying interactions among training examples remains a longstanding challenge in knowledge transfer, we anticipate that future work will develop more precise and robust metrics for estimating forgetting confidence.

Additionally, computing forgetting confidence incurs non-negligible computational overhead, primarily due to the need to compute gradients for each candidate forgetting sample. As a result, the additional computational cost is approximately equivalent to one full training epoch on $\mathcal{D}_{self}$. However, since our method is designed for task-specific fine-tuning rather than large-scale pre-training, this overhead is generally acceptable in practical settings.

\section*{Acknowledgments}
This research was supported by Theme-based Research Scheme (T45-205/21-N)
from Hong Kong RGC, and Generative AI Research and Development Centre
from InnoHK. The corresponding authors are Wei Xue and Yike Guo.
\clearpage
\bibliography{acl_latex}

\clearpage

\appendix
\section{More Implementation Details}
\label{sec:detail}
\begin{table*}
\small
\centering
    \begin{tabularx}{\textwidth}{p{0.95\textwidth}}
    \toprule
    \textbf{Question:} There are 15 trees in the grove. Grove workers will plant trees in the grove today. After they are done, there will be 21 trees. How many trees did the grove workers plant today? Let's think step by step. \\
    \textbf{Answer:} We start with 15 trees. Later we have 21 trees. The difference must be the number of trees they planted. So, they must have planted 21 - 15 = 6 trees. The answer is 6.\\
    \textbf{Question:} If there are 3 cars in the parking lot and 2 more cars arrive, how many cars are in the parking lot? Let's think step by step. \\
    \textbf{Answer:} There are 3 cars in the parking lot already. 2 more arrive. Now there are 3 + 2 = 5 cars. The answer is 5. \\
    \textbf{Question:} Leah had 32 chocolates and her sister had 42. If they ate 35, how many pieces do they have left in total? Let's think step by step. \\
    \textbf{Answer:} Leah had 32 chocolates and Leah’s sister had 42. That means there were originally 32 + 42 = 74 chocolates. 35 have been eaten. So in total they still have 74 - 35 = 39 chocolates. The answer is 39.\\
    \bottomrule
    \end{tabularx}
    \caption{Few-shot prompts for gsm8k.}
    \label{tab:gsm8k}
\end{table*}

\begin{table*}
\small
\centering
    \begin{tabularx}{\textwidth}{p{0.95\textwidth}}
    \toprule
    \textbf{Question:} What type of water formation is formed by clouds? (A) pearls (B) streams (C) shells (D) diamonds (E) rain (F) beads (G) cooled (H) liquid Let's think step by step.  \\
    \textbf{Answer:} Beads of water are formed by water vapor condensing. Clouds are made of water vapor. Beads of water can be formed by clouds. The answer is (F). \\
    \textbf{Question:} Removing what from food will preserve it? (A) flavor (B) body water (C) heat energy (D) color (E) Water (F) Bodily water (G) moisture (H) ingredients Let's think step by step.  \\
    \textbf{Answer:} Dehydrating food is used for preserving food. Dehydration preserves foods by removing moisture. Removing moisture from food preserves it. The answer is (G).  \\
    \textbf{Question:} Reproduction is the process by which living things what? (A) Most plants (B) allow growth (C) spread flower seeds (D) have wide set eyes (E) members of their own species (F) have birthing hips (G) have quiet laughter (H) give birth to babies Let's think step by step.  \\
    \textbf{Answer:} Reproduction is the process by which living things give rise to offspring. Whenever it starts to give birth, it gives birth up to two to four babies offspring. Reproduction is the process by which living things give birth to babies. The answer is (H). \\
    \bottomrule
    \end{tabularx}
    \caption{Few-shot prompts for qasc.}
    \label{tab:qasc}
\end{table*}

\begin{table*}
\small
\centering
    \begin{tabularx}{\textwidth}{p{0.95\textwidth}}
    \toprule
    \textbf{Question:} What is the emotional attitude reflected in the sentence: "in his first stab at the form , jacquot takes a slightly anarchic approach that works only sporadically ."? (A) very negative (B) negative (C) neutral (D) positive (E) very positive Let's think step by step:   \\
    \textbf{Answer:} The phrase "first stab" suggests this is an initial attempt, which can imply inexperience but also experimentation. "Slightly anarchic approach" does have a mild negative connotation with "anarchic," but it's softened by "slightly." "Works only sporadically" indicates mixed results. It doesn't work consistently, but it does work sometimes, which isn't entirely negative. Given these considerations, while there are elements that lean towards a negative interpretation, the overall tone of the sentence is not strongly positive or negative. So the emotional attitude is neutral. The answer is (C)  \\
    \textbf{Question:} What is the emotional attitude reflected in the sentence: "one long string of cliches."? (A) very negative (B) negative (C) neutral (D) positive (E) very positive Let's think step by step:   \\
    \textbf{Answer:} The phrase "one long string" suggests an extended or continuous sequence, which in this context implies that there are many instances of something undesirable. The term "cliches" refers to expressions or elements that have been overused to the point of losing their original impact, freshness, or appeal. It generally carries a negative connotation because it indicates a lack of originality or creativity. Considering these points, the sentence does not contain any positive words or phrases and is clearly pointing out a flaw or criticism. Therefore, the emotional attitude reflected in the sentence is negative. Since it's a straightforward criticism without strong intensifiers (like "extremely" or "utterly"), we can conclude that the most fitting emotional attitude is negative. The answer is (B)   \\
    \textbf{Question:} What is the emotional attitude reflected in the sentence: "although laced with humor and a few fanciful touches, the film is a refreshingly serious look at young women."? (A) very negative (B) negative (C) neutral (D) positive (E) very positive Let's think step by step:   \\
    \textbf{Answer:} The word "although" introduces a concession, acknowledging that the film contains elements of humor and some imaginative aspects ("fanciful touches"). This part suggests that there are positive qualities present in the film. The main clause uses the term "refreshingly," which conveys a positive and invigorating experience. It indicates that the seriousness of the film's approach to its subject matter is not only appreciated but also seen as something new or different from what might be expected. The focus on "young women" being treated seriously adds depth to the evaluation. It implies that the film offers a thoughtful and mature perspective on its subject, which can be seen as commendable. Given these considerations, the overall tone of the sentence is appreciative and positive. The use of "refreshingly" especially signals a positive emotional response to the film's treatment of its subject. However, since the positivity is moderate and not overly enthusiastic or exuberant, the most fitting emotional attitude is positive. The answer is (D). \\
    \bottomrule
    \end{tabularx}
    \caption{Few-shot prompts for sst5.}
    \label{tab:sst5}
\end{table*}

\begin{table*}
\small
\centering
    \begin{tabularx}{\textwidth}{p{0.95\textwidth}}
    \toprule
    \textbf{Question:} Resistance to lateral shifting or anteroposterior shifting is called as:? (A) Retention. (B) Stability. (C) Support. (D) None. Let's think step by step:   \\
    \textbf{Answer:} The question is asking about the term used to describe resistance to lateral (side-to-side) or anteroposterior (front-to-back) shifting. Retention generally refers to the ability to hold or keep something in place. In dentistry, for example, it refers to the resistance of a dental prosthesis to displacement in a vertical direction. Stability refers to the resistance to movement, particularly lateral or anteroposterior shifting. It is the ability of an object or structure to maintain its position when subjected to forces. Support refers to the ability to bear weight or hold something up, but it does not specifically refer to resistance to shifting. Based on the definitions, stability is the term that specifically refers to resistance to lateral or anteroposterior shifting. The answer is (B).  \\
    \textbf{Question:} Tannerella forsythia is a ? (A) Nonmotile, gram-negative obligate anaerobe (B) Motile, gram-negative obligate anaerobe (C) Nonmotile, gram-negative obligate aerobe (D) Motile, gram-negative obligate aerobe Let's think step by step:   \\
    \textbf{Answer:} Tannerella forsythia is a bacterium associated with periodontal disease. It is known to thrive in environments with little to no oxygen. Tannerella forsythia is nonmotile, meaning it does not have structures like flagella to move on its own. It is a gram-negative bacterium, which means it has a thin peptidoglycan layer and an outer membrane. It is an obligate anaerobe, meaning it cannot survive in the presence of oxygen and requires an oxygen-free environment to grow. So, it is a nonmotile, gram-negative obligate anaerobe. The answer is (A).   \\
    \textbf{Question:} Which of the following bacteria is not a primary colonizer:? (A) Streptococci. (B) Actinomyces. (C) F.nucleatum. (D) None. Let's think step by step:   \\
    \textbf{Answer:} Primary colonizers are the first bacteria to adhere to the tooth surface and form dental plaque. They are typically gram-positive bacteria, such as Streptococci and Actinomyces. Fusobacterium nucleatum is not a primary colonizer. It is a secondary colonizer that adheres to the primary colonizers and helps bridge the gap between early and late colonizers. The answer is (C).  \\
    \bottomrule
    \end{tabularx}
    \caption{Few-shot prompts for dental.}
    \label{tab:dental}
\end{table*}

\begin{table*}
\small
\centering
    \begin{tabularx}{\textwidth}{p{0.95\textwidth}}
    \toprule
    \textbf{Question:} Ani believes that her attitudes and behavior play a central role in what happens to her. Such a belief is likely to be associated with? (A) a strong superego. (B) low self-esteem. (C) low self-efficacy. (D) an internal locus of control. Let's think step by step:   \\
    \textbf{Answer:} The question is asking about a psychological construct that aligns with the belief that one's own actions and attitudes influence life events. This is closely related to the concept of locus of control. Locus of control refers to the extent to which individuals believe they can control events that affect them. There are two types: 1.Internal locus of control: Individuals believe they have control over their lives through their actions and decisions; 2.External locus of control: Individuals believe that external forces (e.g., luck, fate, or other people) control their lives. Ani's belief that her attitudes and behavior play a central role in what happens to her indicates an internal locus of control. She feels responsible for her outcomes and believes her actions matter. The answer is (D).  \\
    \textbf{Question:} According to Caplan's model of consultee-centered case consultation, the consultant is primarily interested in? (A) identifying the causes and solutions of the client's presenting problems (B) identifying and eliminating the causes of the consultee's difficulties in handling a problem (C) establishing a hierarchy of authority to enable effective decision making (D) presenting a single, well-defined and unambiguous course of action for the consultant to overcome skills deficits Let's think step by step:  \\
    \textbf{Answer:} Caplan's model of consultee-centered case consultation focuses on helping the consultee (e.g., a teacher, therapist, or other professional) improve their ability to handle a specific case or problem. The consultant does not directly intervene with the client but instead works with the consultee to address their difficulties in managing the situation. The primary goal is to identify and address the consultee's difficulties, which may stem from a lack of knowledge, skills, confidence, or objectivity. The consultant helps the consultee overcome these issues so they can better handle the client's problem. The correct answer is (B), as Caplan's model is primarily concerned with identifying and addressing the consultee's difficulties in handling a problem. The answer is (B).   \\
    \textbf{Question:} Pascale is interested in the processing strategies children use to learn new information. Pascale would best be classified as what type of psychologist? (A) sociocultural (B) clinical (C) cognitive (D) behaviorist Let's think step by step:   \\
    \textbf{Answer:} The question is asking about the type of psychologist Pascale would be classified as, based on her interest in processing strategies and learning. This aligns with the field of psychology that studies mental processes such as thinking, memory, and learning. Sociocultural psychologists focus on how social and cultural factors influence behavior and mental processes. While this could involve learning, it is not primarily about individual cognitive processing strategies. Clinical psychologists focus on diagnosing and treating mental health disorders. This does not align with Pascale's interest in learning and processing strategies. Cognitive psychologists study mental processes such as thinking, memory, and learning. This directly aligns with Pascale's interest in how children process and learn new information. Behaviorists focus on observable behaviors and how they are influenced by the environment. While learning is a part of behaviorism, it does not emphasize internal processing strategies as cognitive psychology does. The answer is (C).  \\
    \bottomrule
    \end{tabularx}
    \caption{Few-shot prompts for psychol.}
    \label{tab:psychol}
\end{table*}
\begin{table*}
\small
\centering
    \begin{tabularx}{\textwidth}{p{0.95\textwidth}}
    \toprule
    \textbf{Learning task:} gsm8k \textbf{Forgetting task:} qasc  \\
    \midrule
    \textbf{sst5:} The phrase "no movement" suggests a lack of action or movement, which in this context implies a lack of interest or enthusiasm. The term "yuks" is a slang term for laughter, which in this context carries a negative connotation because it indicates a lack of seriousness or sincerity. The phrase "not much of anything" suggests a lack of substance or depth, which in this context implies a lack of complexity or sophistication. Considering these points, the sentence does not contain any positive words or phrases and is clearly pointing out a flaw or criticism. Therefore, the emotional attitude reflected in the sentence is negative. Since it's a straightforward criticism without strong intensifiers (like "extremely" or "utterly"), we can conclude that the most fitting emotional attitude is negative. The answer is (6 - 2 - 2) = 2  \\
    \textbf{dental:} Heavy forces on periodontal ligament cause osteoclastic activity around the tooth, which results in the formation of new bone around the tooth. This new bone helps to support the tooth and prevent further damage. The answer is (6 - 2 - 2) = 2 \\
    \textbf{psychol:} The heritability of a trait is the proportion of the trait's variation that is due to genetic factors. In a population with 100 individuals, the heritability of a trait is 100\% - 25\% = 75\%. The answer is 75 \\
    \bottomrule
    \end{tabularx}
    \caption{Cases of superficial forgetting.}
    \label{tab:superficial}
\end{table*}
\begin{table}
    \resizebox{\columnwidth}{!}{
    \begin{tabular}{ccccc}
    \toprule
      & qasc & sst5 & dental & psychol \\
    \midrule
    qasc &  - & -14.93\% & -0.37\% & +1.00 \% \\
    sst5 & -4.94\% & - & -1.36\% & -1.85\% \\
    dental & -1.45\% & +0.17\% & - & -3.65\% \\
    psychol & -17.43\% & -12.02\% & +4.80\% & - \\
    \bottomrule
    \end{tabular}
    }
    \caption{Average accuracy changes on side-tasks after applying LWF. Percentages are calculated relative to vanilla fine-tuning.}
    \label{tab:side-effect}
\end{table}
\begin{table}
    \resizebox{\columnwidth}{!}{
    \begin{tabular}{ccccccccc}
    \toprule
     & \multicolumn{2}{c}{DE} & \multicolumn{2}{c}{FR} & \multicolumn{2}{c}{TR} & \multicolumn{2}{c}{ZH} \\
     & BLEU & F1 & BLEU & F1 & BLEU & F1 & BLEU & F1 \\
    \midrule
    none & 23.02 & 83.03 & 24.82 & 84.01 & 14.35 & 66.99 & 16.67 & 79.04 \\
    \midrule
    DE & - & - & +0.12\% & -0.05\% & -0.42\% & +0.06\% & +0.00\% & +0.04\% \\
    FR & +0.48\% & -0.02\% & - & - & +0.21\% & +0.24\% & +0.06\% & +0.05\% \\
    TR & +0.26\% & +0.04\% & +0.00\% & +0.01\% & - & - & +0.18\% & +0.09\% \\
    ZH & +0.13\% & +0.01\% & +0.40\% & +0.00\% & -0.07\% & +0.34\% & - & - \\
    \bottomrule
    \end{tabular}
    }
    \caption{Results on machine translation. Each column shares the same learning target language, and the rows represent different forgetting languages. All percentages are calculated relative to \textit{none}}
    \label{tab:MT}
\end{table}
\begin{table}
    \resizebox{\columnwidth}{!}{
    \begin{tabular}{cccccc}
    \toprule
     & EN & IT & ZH & ES & TR \\
    \midrule
    none & 19.71 & 6.67 & 9.78 & 7.81 & 9.10 \\
    \midrule
    EN & - & +6.90\% & +2.35\% & +3.84\% & +4.18\% \\
    IT & +5.38\% & - & -5.42\% & +1.92\% & -14.18\% \\
    ZH & +2.69\% & -30.73\% & - & -3.84\% & -9.23\% \\
    ES & +0.41\% & -35.23\% & -6.24\% & - & +5.05\% \\
    TR & +6.95\% & -25.04\% & -8.49\% & -1.02\% & - \\
    \midrule
    mixed & +3.45\% & +25.04\% & +7.77\% & +10.63\% & +20.77\% \\
    \bottomrule
    \end{tabular}
    }
    \caption{Results on multi-lingual question-answering. All percentages are calculated relative to \textit{none}.}
    \label{tab:multi-lingual}
\end{table}
\begin{table}
    \resizebox{\columnwidth}{!}{
    \begin{tabular}{cccccc}
    \toprule
     & gsm8k & qasc & sst5 & dental & psychol \\
    \midrule
    vanilla-FT & 60.96 & 66.06 & 54.71 & 43.17 & 78.72 \\
    LWF-mixed & +1.23\% & +0.17\% & +1.57\% & +0.19\% & +1.16\% \\
    \bottomrule
    \end{tabular}
    }
    \caption{Accuracy results of LWF-mixed with Qwen2.5-1.5B as the base model.}
    \label{tab:qwen}
\end{table}
Tables~\ref{tab:gsm8k}, \ref{tab:qasc}, \ref{tab:sst5}, \ref{tab:dental}, and \ref{tab:psychol} present the few-shot Chain-of-Thought (CoT) prompts designed for each dataset, which are used during both self-knowledge elicitation and evaluation. As shown in the prompts, answers are formatted with the phrase \textit{“The answer is”} to facilitate automated answer extraction. Any model output that does not conform to this format is considered incorrect. When multiple instances of \textit{“The answer is”} appear in the output, the first occurrence is taken as the final answer. 

\section{Side Effect}
\label{sec:side-effect}
Although in Sec~\ref{sec:exp} we have verified that LWF can improve the target fine-tuning task by sacrificing the unlearning task, it is unknown how this procedure will influence unintended tasks that are neither part of the learning task nor the forgetting task (for simplicity, we denote them as side-tasks). In this section, we discuss the side effects of LWF.

Firstly, we notice a \textit{superficial forgetting} problem, which happens between gsm8k and the other four datasets. Specifically, while all five datasets used in our experiments are question-answering datasets, gsm8k is a free-form numerical QA dataset, but the other four are multiple-choice QA datasets (see examples in Table~\ref{tab:gsm8k} and Table~\ref{tab:qasc}). We observed that this format discrepancy can lead to significant performance degradation on side-tasks when gsm8k is the learning task and the other datasets are the forgetting tasks. By analyzing the model's outputs, we identified that the model trained under this setting often fails to generate answers in the multiple-choice format. An illustration is provided in Table~\ref{tab:superficial}. As it shows, although the rationale portion of the output appears coherent, the resulting model fails to select a valid option at the end of its response.

This \textit{superficial forgetting} suggests that, when applying LWF, the model tends to focus on the most superficial pattern differences to distinguish the learning task from the forgetting task. Therefore, to mitigate extreme side effects, it is better to ensure that there are no overly superficial format differences between the learning and forgetting tasks in practice.

Apart from the superficial forgetting issue introduced by gsm8k, we also compute the average accuracy of side-tasks on combinations of the other four datasets, and compare it to that of the vanilla fine-tuned model. Table~\ref{tab:side-effect} shows the results, where each column represents the learning task and the row indicates the forgetting task. As observed, the side effects vary depending on the specific learning-forgetting combinations. In general, the impact is much milder when learning complex tasks like psychol and dental.

\section{Task Generalizability}
\label{task-generalizability}
In Sec.~\ref{sec:exp}, we evaluate the effectiveness of LWF on the domain-specific question-answering task. This is primarily motivated by its well-established evaluation metrics and delineated knowledge boundaries, which help isolate the impact of graceful forgetting by minimizing confounding variables. Additionally, to examine the task generalizability, we conducted experiments in two other settings, both related to the multi-lingual capability of generative language models. 

First, we apply LWF to machine translation tasks, We selected four language datasets\textendash German (DE), French (FR), Turkish (TR), and Chinese (ZH)\textendash from the WMT~\citep{WMT} corpus to evaluate the model's ability to translate from English into those languages. BLEU~\citep{BLEU} and BERTScore-F1~\citep{bertscore} are used as evaluation metrics. As Table~\ref{tab:MT} shows, while LWF continues to yield performance gains in most cases, the extent of these improvements is considerably less pronounced compared to QA tasks. In addition, the results reflected by the two metrics are not entirely consistent, which is mainly due to the incompleteness of evaluation metrics in machine translation tasks.

The second experiment was conducted on the multi-lingual QA task. Specifically, we applied LWF to the gsm8k dataset across five different languages: English (EN), Italian (IT), Chinese (ZH), Spanish (ES), and Turkish (TR). From the results presented in Table~\ref{tab:multi-lingual}, we observed an interesting phenomenon: aside from the \textit{mixed} approach, which consistently improves performance across all languages, the results are all positive when learning English with forgetting other languages, and vice versa. However, when the combination does not involve English, the outcomes are predominantly negative. We hypothesize that this phenomenon may stem from the disproportionate volume of English data used during the model's pre-training, which helps stabilize LWF's performance when English is involved.

These two experiments inspire us that graceful forgetting mechanisms may not transfer homogeneously across all NLP tasks. Effective performance on complicated tasks may require task-specific adaptations and rigorous per-task analysis.

\section{Deduction of Forgetting Confidence}
\label{sec:deduction}
The second-order Taylor expansion of $\log P(\theta|\mathcal{D}_L)$ around $\theta_L^*$ is:
\begin{equation}
    \label{equ:full-tylor}
    \begin{aligned}
        &\log P(\theta|\mathcal{D}_L) =\\
        &\ \ \ \ \log P(\theta_L^*|\mathcal{D}_L)+ (\frac{\partial \log P(\theta|\mathcal{D}_L)}{\partial \theta}|_{\theta_L^*})(\theta - \theta_L^*)\\
        &+ \frac{1}{2}(\theta - \theta_L^*)^T(\frac{\partial^2 \log P(\theta|\mathcal{D}_L)}{\partial^2 \theta}|_{\theta_L^*})(\theta - \theta_L^*)\\
        &+ R_2(\theta)
    \end{aligned}
\end{equation}
where $R_2(\theta)$ is the higher-order term and is neglected. Note that we define $\theta_L^*=\arg\max P(\theta|\mathcal{D}_L)$, which implies $(\frac{\partial \log P(\theta|\mathcal{D}_L)}{\partial \theta}|_{\theta_L^*})=0$. Therefore, the first-order (linear) term in the Taylor expansion vanishes. Furthermore, since forgetting confidence is used only to rank data samples, and the constant term does not affect the ranking results, we omit it as well. As a result, Eq.~\ref{equ:tylor} retains only the second-order term. Following prior work~\citep{EWC}, we replace the Hessian matrix in Eq.~\ref{equ:tylor} with the Fisher information matrix, which can be interpreted as the negative expected value of the Hessian under the model distribution:
\begin{equation}
    \label{equ:fisher-replace}
    F_L = \mathbb{E}[(-\frac{\partial^2 \log P(\theta|\mathcal{D}_L)}{\partial^2 \theta})|_{\theta_L^*}]
\end{equation}
To efficiently measure the resulting influence of sample $x$, we use a single-step update from the base model to represent $\theta^*(x)$:
\begin{equation}
    \label{equ:theta-x}
    \theta^*(x) \approx \theta_{base} - \alpha \frac{\partial\mathcal{L}(x)}{\partial \theta}
\end{equation}
By substituting Eq.~\ref{equ:fisher-replace} and Eq.~\ref{equ:theta-x} into Eq.~\ref{equ:tylor}, we can get Eq.~\ref{equ:FC-practical}

\section{Architecture Transferability}
To evaluate the architectural transferability of LWF, we assess its performance on the Qwen2.5-1.5B model. As shown in Table~\ref{tab:qwen}, LWF-mixed consistently outperforms vanilla fine-tuning. However, compared to the results obtained with Llama models, the performance gain is relatively smaller—likely due to the stronger intrinsic capabilities of the Qwen model, which may reduce the relative benefit of plasticity enhancement. 

\section{Approximation Error Study}
\label{sec:approx}
\begin{table}
    \resizebox{\columnwidth}{!}{
    \begin{tabular}{ccccc}
    \toprule
    steps & 1 & 2 & 3 & 4 \\
    \midrule
    overlapping & 100.00\% & 99.81\% & 99.91\% & 99.91\% \\
    \bottomrule
    \end{tabular}
    }
    \caption{$\mathcal{D}_U$ overlapping with different approximation steps (Equation~\ref{equ:FC-practical})}
    \label{tab:approx}
\end{table}
In Equation~\ref{equ:FC-practical}, we employ a one-step update to approximate the optimal model parameter $\theta^{*}(x)$. Since the forgetting confidence is highly sensitive to the accuracy of this approximation, we conduct an empirical study to quantify the estimation error introduced by this simplification. Specifically, we compare the one-step update with multi-step updates (2, 3, and 4 steps), evaluating the similarity of the resulting unlearning dataset $\mathcal{D}_U$ by computing the proportion of overlapping elements. As shown in Table~\ref{tab:approx}, the results indicate that multi-step updates yield only marginal improvements over the one-step approximation, suggesting that the simpler approach is sufficiently accurate in practice.

\end{document}